\documentclass{article}
\usepackage{spconf,amsmath,amssymb,amsfonts,graphicx}
\usepackage{booktabs}
\usepackage{placeins}

\usepackage{xcolor}
\usepackage[pagebackref=false,breaklinks=true,colorlinks,bookmarks=false]{hyperref}

\title{SCOPE-AD: SEQUENTIAL COST-AWARE ORDINAL-BELIEF PLANNING WITH ENERGY-BASED MODELS FOR DIAGNOSTIC AGENTS}

\name{\begin{tabular}{@{}c@{}}Ziwen Yu$^{1}$, Ivan Koychev$^{2}$, Elizabeth Coulthard$^{1}$, Ting Zhou$^{1}$, Bolin Chen$^{1}$,\\
 Dian Hong$^{4}$, Zinuo You$^{1}$, Yujiao Wang$^{3}$, Anthony Mulholland$^{1}$, Qiang Liu$^{1}$\end{tabular}}
\address{$^{1}$University of Bristol, Bristol, UK\\
$^{2}$Imperial College London, London, UK\\
$^{3}$Manchester Metropolitan University, Manchester, UK\\ $^{4}$Hunan Normal University, Changsha, China}

\begin{document}
\ninept
\maketitle

\begin{abstract}
Alzheimer's disease (AD) diagnosis requires sequential evidence acquisition under heterogeneous test costs and patient burden. Fixed-modality predictors do not jointly decide which test to acquire or when the available evidence is sufficient for diagnosis. We propose SCOPE-AD (Sequential Cost-Aware Ordinal-Belief Planning with Energy-Based Models for Diagnostic Agents) for cost-aware classification of cognitively normal (CN), mild cognitive impairment (MCI), and AD cases. A mask-aware ordinal model represents uncertainty along the ordered CN--MCI--AD continuum. Retrospective training records provide sampled Bellman targets for an energy-based teacher, whose action distributions are distilled into a Qwen policy. At deployment, the agent selects acquisition or diagnosis actions under availability and budget constraints without access to unacquired values. After each acquisition, the evidence and ordinal belief are updated before the next decision. On ADNI, SCOPE-AD achieves 77.70\% Macro-F1 at an average acquisition cost of \$50.46, exceeding the strongest evaluated baseline by 9.34 percentage points. Full-modality evaluation raises Macro-F1 by only 1.89 points while increasing acquisition cost by 116.7 times. These results support selective acquisition for cost-effective diagnosis.

\end{abstract}


\begin{figure*}[!t]
    \centering
    \includegraphics[width=0.93\linewidth]{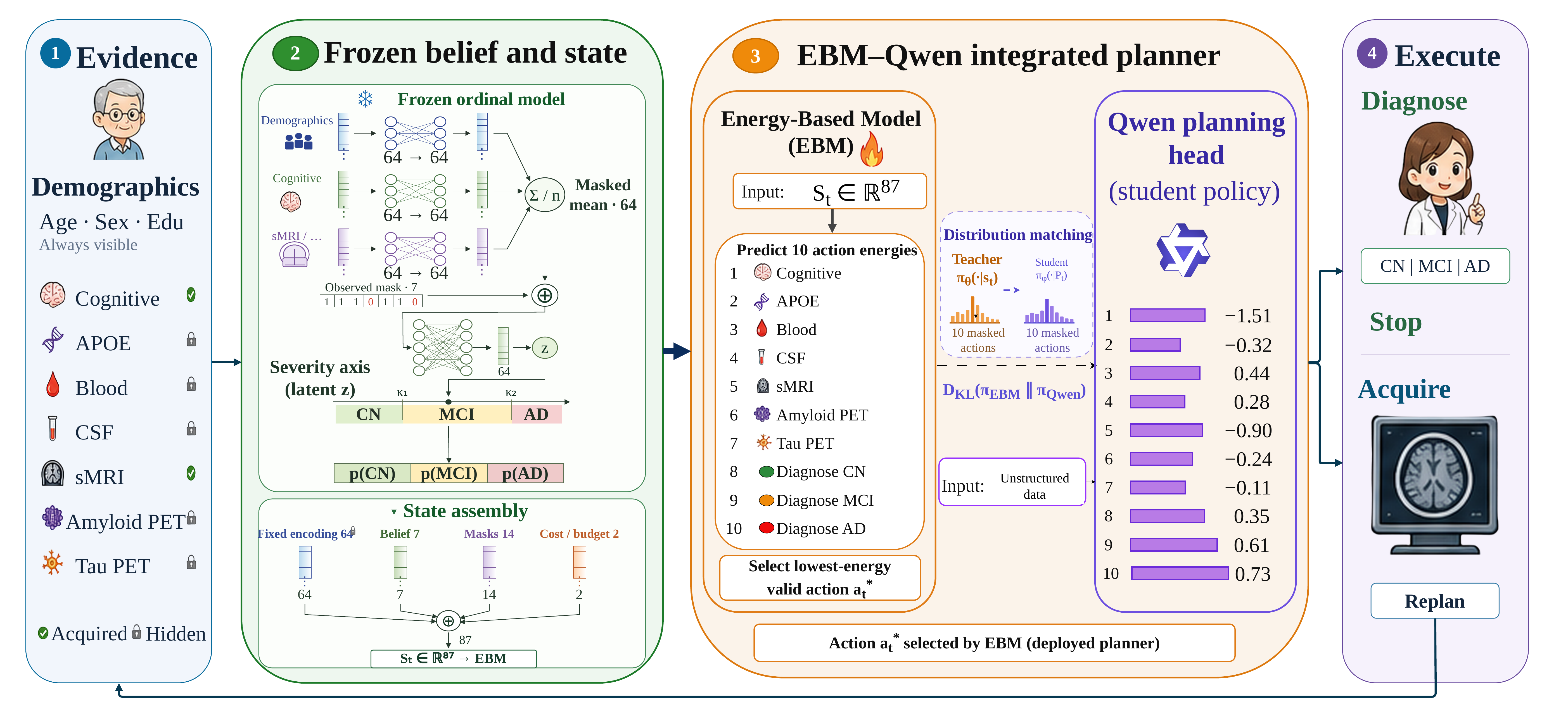}
    \caption{\textbf{Overview of SCOPE-AD.} Calibrated ordinal beliefs and the current acquisition context form the planner state. A finite-horizon energy-based model (EBM) planner uses soft Bellman backups to evaluate modality-acquisition and terminal diagnosis actions under cost and feasibility constraints. The planner’s policy is distilled into a Qwen agent conditioned on structured prompts. After each acquisition, the state is updated and the Qwen policy selects the next action until a terminal diagnosis is reached.}
    \label{fig:example}
\end{figure*}

\section{Introduction}

Alzheimer's disease (AD) is a progressive neurodegenerative disease, with clinical status commonly characterized as cognitively normal (CN), mild cognitive impairment (MCI), and AD along an ordered severity continuum~\cite{jack2024revised,pichet2025evaluation}. Accurate discrimination among these clinical states is important for timely diagnosis and disease management. In practice, AD diagnosis is not necessarily a one-shot prediction problem, but often a sequential evidence-acquisition process in which decisions are made from partially observed patient information~\cite{grande2025blood}. Clinicians must repeatedly determine whether the currently available evidence is sufficient for diagnosis or whether an additional test should be acquired, while candidate tests can differ substantially in monetary cost, invasiveness, and patient burden~\cite{frisoni2025new}. This gives rise to a central decision problem in AD diagnosis: when should the current evidence be considered sufficient, and, if it is not, which additional diagnostic modality is worth acquiring next?

Existing work on sequential diagnosis spans several methodological directions. Benchmarks such as MIMIC-CDM, MediQ, and AgentClinic evaluate interactive information seeking under incomplete clinical evidence~\cite{schmidgall2025agentclinicmultimodalagentbenchmark,hager2024evaluation,li2024mediq}, whereas systems such as AMIE and MAI-DxO use self-play, dialogue fine-tuning, or multi-agent orchestration to support iterative diagnosis~\cite{tu2025towards,nori_sequential_2025}. A second line learns test-selection policies more directly: LA-CDM and DiagAgent optimize large language model (LLM) agents through supervised and reinforcement learning (RL), while ED-Copilot combines language-model representations with an RL controller~\cite{bani2026language,qiu2025evolving,sun2024edcopilot}. Other approaches replace reward-driven policy learning with explicit probabilistic criteria; ACTMED applies Bayesian Experimental Design to select tests with the greatest expected reduction in diagnostic uncertainty~\cite{estevez2025timely}. In parallel, non-LLM active feature acquisition methods formulate diagnostic testing as a Markov decision process (MDP) and learn cost-sensitive policies through deep, batch, or model-based RL~\cite{zhu2021learning,yu2023deep}. Despite these advances, most methods target differential diagnosis among alternative disease hypotheses and do not explicitly represent uncertainty along an ordered disease continuum.

This limitation is particularly relevant to AD diagnosis. Artificial intelligence (AI)-based AD studies remain dominated by binary tasks such as AD--CN, MCI--CN, and AD--MCI, while multimodal models commonly assume that all required MRI, PET, CSF, or clinical measurements are available before prediction~\cite{frisoni2025new, jack2024revised, yu2026multimodal}. Ordinal models explicitly encode label order, unlike conventional multiclass models, but encoding this order alone does not determine which modality to acquire or when to stop~\cite{pichet2025evaluation}. Conversely, recent acquisition-aware approaches, including UMoS, uncertainty-based MRI-to-PET referral, ProMUSE, and AD-specific RL policies, provide only partial solutions: they focus on conversion prediction or binary classification, use restricted modality sequences, or learn actions from categorical patient states~\cite{doan2026promuseprogressivemultimodaluncertaintyguided,zheng2024uncertainty}. Consequently, an open problem is how to represent diagnostic uncertainty along the ordered CN--MCI--AD continuum from partially observed multimodal evidence, and how to place terminal diagnoses and acquisition trajectories in a unified decision space where diagnostic risk can be compared with the cost-adjusted value of further acquisition.

To address this gap, we propose \textbf{SCOPE-AD}
(Sequential Cost-Aware Ordinal-Belief Planning with Energy-Based Models for Diagnostic Agents), which uses ordinal beliefs to guide modality acquisition and diagnostic stopping under cost and feasibility constraints. Fig.~\ref{fig:example} provides an overview. Our contributions are:
\begin{itemize}
\item We formulate CN--MCI--AD diagnosis as cost-aware sequential decision-making with ordinal beliefs and acquisition and terminal-diagnosis actions.
\item We develop SCOPE-AD, combining cross-fitted ordinal beliefs, an offline-trained EBM teacher, and policy distillation into a deployed Qwen agent.
\item We construct a sequential ADNI evaluation protocol, where SCOPE-AD achieves 77.70\% Macro-F1 at an average acquisition cost of \$50.46.
\end{itemize}

\section{Proposed Method}
\label{sec:method}

\subsection{Overview}

SCOPE-AD constructs ordinal beliefs from observed evidence, trains an offline EBM teacher on retrospective Bellman targets, and distills its policy into Qwen for sequential acquisition and diagnosis.

\subsection{Ordinal Belief Model}

For participant $i$ at decision step $t$, let $\mathbf x_i^{(0)}$ be always-visible information and $\mathbf x_i^{(m)}$ features of requestable modality $m\in\mathcal M$. The observation mask $\mathbf o_{i,t}=[o_{i,t}^{(m)}]_{m\in\mathcal M}\in\{0,1\}^{|\mathcal M|}$ uses 1 for acquired modalities. Independent encoders $\phi_0,\phi_m$ embed these inputs; masked fusion and the ordinal head $f_{\mathrm{ord}}$ yield severity $\eta_{i,t}$. Ordered thresholds at boundaries $\mathcal B=\{\mathrm{CN},\mathrm{MCI}\}$ and temperature calibration give the CN--MCI--AD posterior:

\begin{equation}
\begin{aligned}
\bar{\mathbf e}_{i,t}
&=
\frac{
\phi_0(\mathbf x_i^{(0)})
+\sum_{m\in\mathcal M}
o_{i,t}^{(m)}\phi_m(\mathbf x_i^{(m)})
}{
1+\sum_{m\in\mathcal M}o_{i,t}^{(m)}
},\\
\eta_{i,t}
&=
f_{\mathrm{ord}}([\bar{\mathbf e}_{i,t};\mathbf o_{i,t}]),\\
\kappa_{\mathrm{MCI}}
&=
\kappa_{\mathrm{CN}}+\operatorname{softplus}(\rho)+\epsilon,\\
F_{i,t}^{(b)}
&=
\sigma\!\left(
\frac{\kappa_b-\eta_{i,t}}{T_{\mathrm{cal}}}
\right),\quad b\in\mathcal B,\\
\mathbf p_{i,t}
&=
\left[
F_{i,t}^{(\mathrm{CN})},
F_{i,t}^{(\mathrm{MCI})}-F_{i,t}^{(\mathrm{CN})},
1-F_{i,t}^{(\mathrm{MCI})}
\right].
\end{aligned}
\label{eq:ordinal_belief}
\end{equation}

Here, $\bar{\mathbf e}_{i,t}$ denotes the fused embedding, $\kappa_b$ is the threshold for boundary $b$, and $F_{i,t}^{(b)}$ is the corresponding cumulative probability. The ordinal model learns parameters $\psi$ by minimizing the negative log-likelihood of the ground-truth diagnosis:
\begin{equation}
\mathcal L_{\mathrm{ordinal}}(\psi)
=
-\frac{1}{N}\sum_{i=1}^{N}
\log p_{\psi,i}^{y_i},
\label{eq:ordinal_loss}
\end{equation}
where $N$ counts training samples and $p_{\psi,i}^{y_i}$ is the posterior probability of ground-truth diagnosis $y_i$.

Here, $\rho$ is an unconstrained threshold-gap parameter, $\epsilon>0$ ensures strict ordering, $\sigma$ is the logistic sigmoid, and $T_{\mathrm{cal}}>0$ is the calibration temperature. Patient-level cross-fitting provides out-of-fold training beliefs; a fixed feature-hashing encoder gives fold-consistent observed-feature representations $\mathbf h_{i,t}$. The state includes posterior entropy $H_{i,t}=-\sum_{y\in\mathcal Y}p_{i,t}^{y}\log p_{i,t}^{y}$, expected ordinal rank $\mu_{i,t}=\sum_{y\in\mathcal Y}r(y)p_{i,t}^{y}$, two boundary-evidence scores $\boldsymbol\delta_{i,t}$, and normalized accumulated cost $\widetilde C_{i,t}$ and remaining budget $\widetilde B_{i,t}$. The availability mask $\mathbf u_i$ identifies requestable modalities before any action without revealing outcomes. The planner state is $\mathbf s_{i,t}=[\mathbf h_{i,t};\mathbf p_{i,t};H_{i,t};\mu_{i,t};\boldsymbol\delta_{i,t};\mathbf o_{i,t};\mathbf u_i;\widetilde C_{i,t};\widetilde B_{i,t}]$, with belief $p(y\mid\mathbf s_{i,t}):=p_{i,t}^{y}$ for $y\in\mathcal Y=\{\mathrm{CN},\mathrm{MCI},\mathrm{AD}\}$.

\subsection{Offline Energy-Based Planning}

Actions request modalities or terminate with diagnosis $k$. The hard valid-action mask $M(\mathbf s,a)\in\{0,1\}$ excludes unavailable, observed, over-budget, or beyond-horizon requests; all acquisition methods mask unavailable modalities. Requesting $m$ incurs normalized cost penalty $g_m$; terminal risk $R_k(\mathbf s)$ combines expected classification error and ordinal distance. Training transition $\mathcal T_i(\mathbf s,m)$ reveals participant $i$'s recorded modality outcome and updates the state, yielding record-specific sampled Bellman targets for energy predictor $E_\theta$:

\begin{equation}
\begin{aligned}
R_k(\mathbf s)
&=
\sum_{y\in\mathcal Y}p(y\mid\mathbf s)
\left[
\lambda_{\mathrm{cls}}\mathbb I(y\neq k)
+\lambda_{\mathrm{ord}}|r(y)-r(k)|
\right],\\
\widehat E_{i,d}(\mathbf s,a)
&=
\begin{cases}
R_k(\mathbf s), & a=\textsc{Diagnose}_k,\\
g_m+\gamma\widehat V_{i,d-1}(\mathcal T_i(\mathbf s,m)),
& a=\textsc{Request}_m,
\end{cases}\\
\widehat V_{i,d}(\mathbf s)
&=
-\tau_E\log
\sum_{a:M(\mathbf s,a)=1}
\exp[-\widehat E_{i,d}(\mathbf s,a)/\tau_E],\\
\pi_{\theta}(a\mid\mathbf s)
&=
\frac{
M(\mathbf s,a)\exp[-E_{\theta}(\mathbf s,a)/\tau_E]
}{
\sum_{a'}M(\mathbf s,a')
\exp[-E_{\theta}(\mathbf s,a')/\tau_E]
},\\
\mathcal L_{\mathrm{EBM}}
&=
\operatorname{Huber}_{M}(E_{\theta},\widehat E_{i,d})
+\lambda_{\mathrm{rank}}
D_{\mathrm{KL}}(\widehat\pi_{i,d}\,\|\,\pi_{\theta}).
\end{aligned}
\label{eq:energy_planner}
\end{equation}

Here, $\theta$ denotes EBM parameters, $\mathbb I$ the indicator function, and $\lambda_{\mathrm{cls}},\lambda_{\mathrm{ord}},\lambda_{\mathrm{rank}}$ weight classification error, ordinal distance, and policy matching, respectively. The targets $\widehat E_{i,d}$ and $\widehat V_{i,d}$ are sampled action energies and soft state values. The function $r(\cdot)$ gives diagnosis rank; $\gamma$ is the continuation factor and $\tau_E>0$ the soft-backup temperature. For horizon $D$ and one acquisition per step, remaining depth $d=D-\sum_{m\in\mathcal M}o_{i,t}^{(m)}$ is determined by the state mask; $d=0$ permits only terminal diagnoses. The target $\widehat\pi_{i,d}$ is the masked soft policy induced by $\widehat E_{i,d}$. Masked Huber loss fits valid-action energies; $D_{\mathrm{KL}}$ denotes Kullback--Leibler (KL) divergence for policy matching. The fitted $E_\theta$ uses only the current state to provide the distillation teacher policy $\pi_\theta$.

\subsection{Planner-to-Agent Policy Distillation}

EBM rollouts on training records reveal outcomes only after requests, forming $\mathcal D_{\mathrm{teacher}}$ of states, valid-action masks, and soft policies. The multilayer perceptron (MLP) student receives numerical states; Qwen receives structured prompts encoding observed information, belief, availability, costs, budget, and valid actions. Both student policies $\pi_\varphi$ match the masked teacher policy:

\begin{equation}
\begin{aligned}
\pi_{\varphi}(a\mid\mathbf s)
&=
\frac{
M(\mathbf s,a)\exp[\ell_{\varphi}(\mathbf s,a)/\tau_D]
}{
\sum_{a'}M(\mathbf s,a')
\exp[\ell_{\varphi}(\mathbf s,a')/\tau_D]
},\\
\mathcal L_{\mathrm{distill}}
&=
\mathbb E_{\mathbf s\sim\mathcal D_{\mathrm{teacher}}}
\left[
D_{\mathrm{KL}}
\left(
\pi_{\theta}(\cdot\mid\mathbf s)
\,\|\,\pi_{\varphi}(\cdot\mid\mathbf s)
\right)
\right].
\end{aligned}
\label{eq:agent_distillation}
\end{equation}

Here, $\varphi$ denotes student parameters, $\ell_\varphi$ action logits, and $\tau_D>0$ the distillation temperature. At deployment, Qwen acts under the hard mask without EBM calls or unacquired values. Acquisitions update evidence, accumulated cost, remaining budget, and the observation mask, then recompute the ordinal belief and state; terminal diagnoses end the sequence.

\section{Experiments and Analysis}
\label{sec:exp}

On the ADNI benchmark, we compared SCOPE-AD with LA-CDM~\cite{bani2026language}, ReAct~\cite{yao2023reactsynergizingreasoningacting}, ACTMED~\cite{estevez2025timely}, SM-DDPO~\cite{yu2023deep}, and LLM baselines MedGemma-27B and Qwen2.5-32B.

\subsection{Dataset}

\textbf{ADNI.} ADNI provides multimodal clinical and biomarker data for AD research~\cite{weiner2017alzheimer,jack2008alzheimer}. We constructed a retrospective sequential-acquisition benchmark using one baseline record per participant. We retained 2,218 participants with valid CN, MCI, or AD labels and used a subject-level split of 1,552/333/333 training, validation, and test cases. Demographics were initially visible, while the seven modality groups in Tab.~\ref{tab:adni_modalities} were requestable. The six cognitive features were selected to exclude variables used to define the CN, MCI, and AD labels in our benchmark. Patient-level modality availability was known before the first action, but recorded modality values remained hidden until the corresponding request. Unavailable modalities were excluded through hard action masking. Each modality was assigned a prespecified acquisition cost in U.S. dollars, defining the benchmark's performance--cost setting.

\begin{table}[t]
\centering
\caption{Requestable ADNI modalities, input-feature dimensions, test-set availability, and acquisition costs. The total cost assumes that all seven modalities are available.}
\label{tab:adni_modalities}
\setlength{\tabcolsep}{4pt}
\begin{tabular}{lrrr}
\toprule
Modality &
Input feature dims. &
Test availability &
Cost \\
\midrule
Cognitive scales & 6  & 333 (100.0\%) & \$50 \\
APOE genotype    & 7  & 323 (97.0\%)  & \$283 \\
Blood biomarkers & 39 & 144 (43.2\%)  & \$232 \\
CSF biomarkers   & 8  & 241 (72.4\%)  & \$574 \\
Structural MRI   & 16 & 333 (100.0\%) & \$1,431 \\
Amyloid PET      & 28 & 264 (79.3\%)  & \$2,975 \\
Tau PET          & 14 & 141 (42.3\%)  & \$2,975 \\
\midrule
Total            & 118 & --- & \$8,520 \\
\bottomrule
\end{tabular}
\end{table}

\subsection{Experiment Setup}

The ordinal belief model used two-layer 64-unit modality encoders, a 64-unit fusion layer, 0.20 dropout, temperature calibration, and five-fold patient-level cross-fitting, yielding an 87-dimensional planner state. The EBM planner was a 256--256--128 Gaussian error linear unit (GELU) MLP with 0.10 dropout, trained using AdamW (learning rate $10^{-3}$, weight decay $10^{-4}$, and batch size 512) with planning horizon $D=7$, $\gamma=1$, and $\tau_E=0.3$. The Qwen3.5-9B~\cite{qwen3.5} agent used quantized low-rank adaptation (QLoRA) with 4-bit NormalFloat (NF4) quantization~\cite{dettmers2023qlora} ($r=8$, $\alpha=16$, dropout 0.05) and soft-target KL distillation with AdamW (learning rate $2\times10^{-4}$ and effective batch size 16). Results are reported as the mean over three runs with seeds 42, 123, and 2026. We report class-wise recall for CN, MCI, and AD, the macro-averaged F1 score (Macro-F1), and average test cost per patient. During evaluation, only successfully acquired modalities contributed to cost.

\begin{table}[t]
\centering
\caption{Evaluation results on ADNI.}
\label{tab:adni_evaluation_results2}
\setlength{\tabcolsep}{1.5pt}
\begin{tabular}{@{}lccccc@{}}
\toprule
Method & \multicolumn{3}{c}{\shortstack{Recall\\[-2pt](\%)}} & \shortstack{Macro-F1\\[-2pt](\%)} & \shortstack{Avg. test\\[-2pt]cost} \\
\cmidrule(lr){2-4}
& CN & MCI & AD & & \\
\midrule
LA-CDM~\cite{bani2026language}
& 71.39 & 48.51 & 60.99 & 60.32 & \$794.33 \\
MedGemma-27B
& \textbf{94.56} & 55.63 & 17.73 & 56.18 & \textbf{\$49.77} \\
Qwen2.5-32B
& 74.70 & 24.14 & 64.54 & 46.76 & \$4,784.50 \\
SM-DDPO~\cite{yu2023deep}
& 78.01 & 59.31 & 68.79 & 67.63 & \$138.74 \\
ReAct~\cite{yao2023reactsynergizingreasoningacting}
& 62.65 & 34.25 & \textbf{75.18} & 50.38 & \$2,091.60 \\
ACTMED~\cite{estevez2025timely}
& 79.43 & 62.53 & 64.54 & 68.36 & \$3,171.72 \\
\midrule
\textbf{SCOPE-AD (Ours)}
& 92.67 & \textbf{63.91} & \textbf{75.18}
& \textbf{77.70} & \$50.46 \\
\bottomrule
\end{tabular}
\end{table}

\subsection{Comparison with State-of-the-Art Methods}

Tab.~\ref{tab:adni_evaluation_results2} compares the deployed Qwen policy with language-model and active acquisition baselines. SCOPE-AD achieves the highest Macro-F1 of 77.70\%, exceeding ACTMED, the strongest baseline on this metric, by 9.34 percentage points. It also achieves the highest MCI recall of 63.91\% and ties with ReAct for the highest AD recall of 75.18\%, while its CN recall of 92.67\% is second to MedGemma.

The average acquisition cost is \$50.46, only \$0.69 above MedGemma and 63.6\% and 98.4\% below SM-DDPO and ACTMED, respectively. These results indicate a favorable trade-off between diagnostic performance and acquisition cost under the evaluated protocol and prespecified costs. Compared with the strongest active-acquisition baselines, SCOPE-AD achieves higher Macro-F1 at markedly lower cost, indicating more efficient evidence acquisition.

\begin{figure}[!t]
    \centering
    \includegraphics[width=\columnwidth]{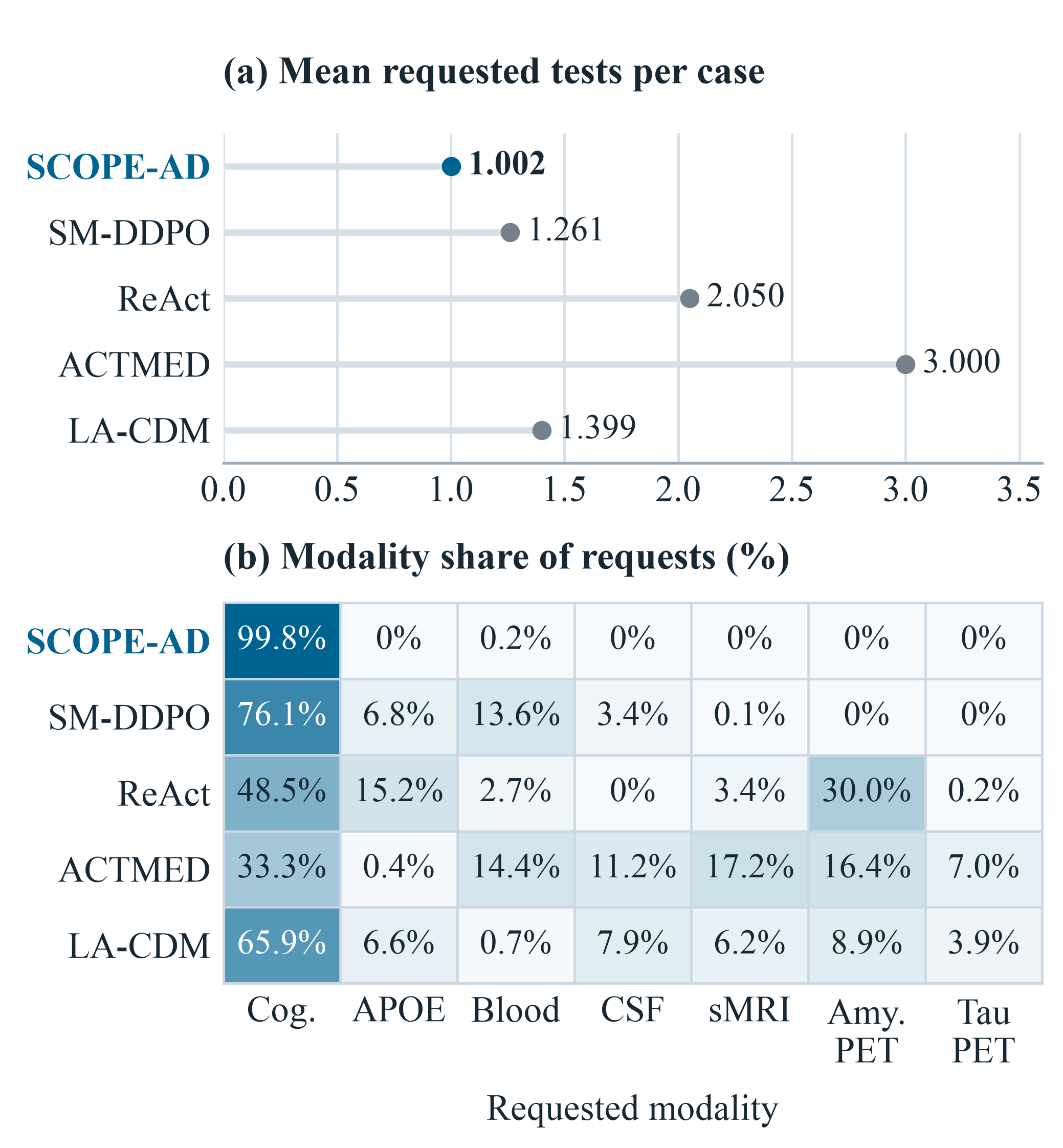}
    \caption{Acquisition frequency and modality composition on ADNI. (a) Mean number of modality requests per case. Error bars indicate variability when available. (b) Percentage of requests per modality.}
    \label{fig:acquisition_behavior}
\end{figure}

Fig.~\ref{fig:acquisition_behavior} shows that SCOPE-AD makes approximately one modality request per case, with cognitive assessments accounting for 99.8\%
of all requests. The policy thus usually acquires cognitive evidence and stops without requesting expensive modalities.

\subsection{Ablation and Cost--Utility Analysis}

\begin{table}[!htbp]
\centering
\caption{Ablations of SCOPE-AD. Bold marks the highest recall/Macro-F1 and lowest cost, including ties.}
\label{tab:ablation}
\setlength{\tabcolsep}{1.5pt}
\begin{tabular}{@{}lccccc@{}}
\toprule
Variant & \multicolumn{3}{c}{\shortstack{Recall\\[-2pt](\%)}} & \shortstack{Macro-F1\\[-2pt](\%)} & \shortstack{Avg. test\\[-2pt]cost} \\
\cmidrule(lr){2-4}
& CN & MCI & AD & & \\
\midrule
w/o Ordinal Belief
& 78.72 & {60.69} & {68.79} & 67.20 & \textbf{\$50.00} \\
w/o EBM Planner
& 66.67 & 52.87 & 62.41 & 61.10 & {\$120.50} \\
\midrule
SCOPE-AD (Full-modality)
& 82.98 & \textbf{70.34} & \textbf{78.72} & \textbf{79.59} & \$5,889.49 \\
Cognitive Only
& 85.11 & 59.31 & \textbf{78.72} & 74.34 & \textbf{\$50.00} \\
w/o Cognitive
& 78.25 & 68.28 & 40.43 & 64.60 & \$1,839.49 \\
\textbf{SCOPE-AD (ours)}
& \textbf{92.67} & 63.91 & 75.18 & 77.70 & \$50.46 \\
\bottomrule
\end{tabular}
\end{table}

Tab.~\ref{tab:ablation} reports component ablations and fixed-policy modality controls. The w/o Ordinal Belief and w/o EBM Planner variants achieve Macro-F1 scores of 67.20\% and 61.10\%, respectively, compared with 77.70\% for SCOPE-AD.

The modality controls reuse the same trained Qwen checkpoint without further fine-tuning and jointly characterize the cost--utility structure learned by the policy. Cognitive Only starts from demographics and permits only cognitive acquisition and terminal diagnosis actions, whereas the w/o Cognitive condition disables cognitive requests and forces the policy to select among the remaining modalities. Full-Modality reveals all available modalities and charges their corresponding acquisition costs, providing an empirical performance ceiling for the same deployed policy under the benchmark.

Cognitive Only achieves 74.34\% Macro-F1 at \$50.00, showing that cognitive evidence alone provides high diagnostic utility at very low acquisition cost in this ADNI benchmark. SCOPE-AD improves this result to 77.70\% at only \$50.46, while its learned policy assigns 99.8\% of acquisition requests to cognitive assessments. In contrast, when cognitive acquisition is disabled, the same policy redirects acquisition toward alternative modalities and retains 64.60\% Macro-F1, although at a substantially higher average cost of \$1,839.49. This indicates that the policy is not restricted to a fixed cognitive-only acquisition rule, but can exploit non-cognitive evidence when the preferred modality is unavailable.

Full-Modality reaches 79.59\% Macro-F1 at \$5,889.49. Thus, revealing all available modalities improves Macro-F1 by only 1.89 percentage points over SCOPE-AD while increasing acquisition cost to 116.7 times that of SCOPE-AD. Taken together, these controls suggest that the predominance of cognitive acquisition reflects the cost--utility structure of the benchmark rather than an inability to use other modalities: cognitive assessments provide most of the attainable diagnostic benefit at low cost, whereas additional modalities offer comparatively limited marginal diagnostic gain relative to their acquisition cost. These findings support selective acquisition over full-modality use. The learned policy therefore tends to stop after acquiring cognitive evidence when its expected marginal benefit from further testing does not justify the additional cost, while retaining the ability to follow alternative acquisition trajectories when necessary.

\FloatBarrier
\section{Conclusion}
\label{sec:con}

In this work, we proposed SCOPE-AD, a cost-aware ordinal belief--energy planning framework for sequential CN--MCI--AD diagnosis from partially observed multimodal evidence. By integrating ordinal belief estimation, EBM-based planning, and planner-to-agent policy distillation, SCOPE-AD jointly determines whether to acquire evidence or make a terminal diagnosis. Experiments on the ADNI-based sequential diagnostic benchmark show that SCOPE-AD achieves 77.70\% Macro-F1 at an average cost of \$50.46, while full-modality acquisition provides only a 1.89-point additional gain at substantially higher cost. These results demonstrate that SCOPE-AD balances diagnostic performance and acquisition cost by avoiding tests with limited marginal diagnostic benefit. The ablation and cost--utility analyses further indicate that both ordinal uncertainty modeling and unified acquisition--diagnosis planning contribute to this favorable performance--cost trade-off. Overall, SCOPE-AD provides an effective framework for cost-aware sequential diagnosis under partial multimodal evidence. Future work will extend to longitudinal patient records and dynamic disease progression. Code available: \url{https://github.com/chijiaozima/SCOPE-AD}.
\clearpage
\bibliographystyle{IEEEbib}
\bibliography{refs}

\end{document}